\documentclass{article}

\usepackage{microtype}
\usepackage{graphicx}
\usepackage{subfigure}
\usepackage{booktabs} 

\usepackage{hyperref}

\usepackage[accepted]{icml2025}

\usepackage{amsmath}
\usepackage{amssymb}
\usepackage{mathtools}
\usepackage{amsthm}
\usepackage{makecell}
\usepackage[textsize=tiny]{todonotes}
\usepackage{pifont}
\usepackage{colortbl} 

\usepackage[capitalize,noabbrev]{cleveref}

\theoremstyle{plain}

\theoremstyle{definition}

\theoremstyle{remark}

\definecolor{wkred}{RGB}{255, 190, 190}
\definecolor{wkblue}{RGB}{210, 230, 250}

\icmltitlerunning{Submission and Formatting Instructions for ICML 2025}

\begin{document}

\twocolumn[
\icmltitle{EA4LLM: A Gradient-Free Approach to Large Language Model Optimization via Evolutionary Algorithms}




\begin{icmlauthorlist}
\icmlauthor{WenTao Liu}{11,33}
\icmlauthor{Siyu Song}{22}
\icmlauthor{Hao Hao}{11}
\icmlauthor{Aimin Zhou}{11,33}


\end{icmlauthorlist}

\icmlaffiliation{11}{Shanghai Institute of Artifical Intelligence Education, East China Normal University, Shanghai, China}
\icmlaffiliation{22}{School of Computer Science and Technology, East China Normal University, Shanghai, China}
\icmlaffiliation{33}{Shanghai Innovation Institute, Shanghai, China}


\icmlcorrespondingauthor{Aimin Zhou}{amzhou@cs.ecnu.edu.cn}

\icmlkeywords{Machine Learning, ICML}

\vskip 0.3in
]



\printAffiliationsAndNotice{}  

\begin{figure*}[t!]
  \vspace{-2mm}
  \centering
  \includegraphics[width=0.8\textwidth]{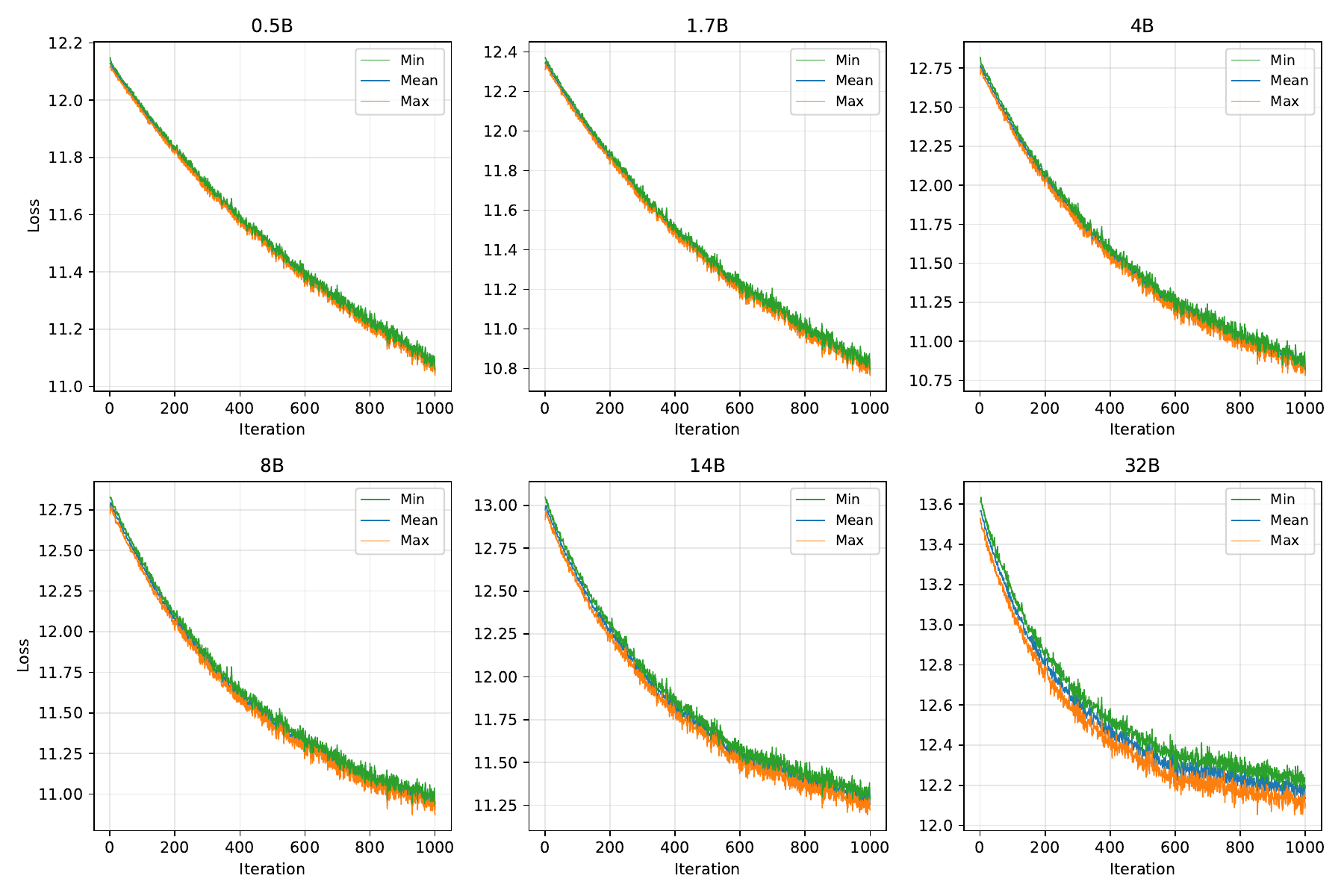}
  \vspace{-6mm}
  \caption{Loss Curve of EAGPT Optimized from Pretraining Using Evolutionary Algorithms.}
  \label{fig:fig1_loss}
  \vspace{-3mm}
\end{figure*}

\begin{abstract}
In recent years, large language models (LLMs) have made remarkable progress, with model optimization primarily relying on gradient-based optimizers such as Adam.  
However, these gradient-based methods impose stringent hardware requirements, demanding high-concurrency, high-memory GPUs. Moreover, they require all neural network operations to be differentiable, thereby excluding many promising non-differentiable architectures from practical use.

To address these limitations, we propose EA4LLM, an evolutionary algorithm for optimizing LLMs, and, for the first time, empirically verify full-parameter optimization from the pretraining stage across model sizes ranging from 0.5B to 32B. We conduct extensive experiments and provide key insights into how evolutionary algorithms can effectively optimize neural networks.

Our work challenges the prevailing assumption that gradient-based optimization is the only viable approach for training neural networks. It also holds significant potential to reduce the computational cost of training large language models, thereby enabling groups with limited computational resources to participate in deep learning research.
\end{abstract}

\section{Introduction}
\label{sec:intro}

Over the past few decades, gradient-based optimization has been the dominant—and often the main approach for training neural networks. A series of improved gradient-based algorithms have been successively proposed, including Stochastic Gradient Descent (SGD)~\cite{robbins1951stochastic}, Adaptive Moment Estimation (Adam)~\cite{kingma2014adam}, and more recently, EvoLved Sign Momentum (Lion)~\cite{chen2023lion}.

However, gradient-based optimization suffers from two major limitations. First, it requires computing and storing gradients during training, leading to substantial memory and computational overhead. In practice, training typically incurs 3–8$\times$ the model parameter size in GPU memory (for parameters, gradients, and optimizer states) and 2–3$\times$ the computational cost of a forward pass (due to forward propagation, backpropagation, and gradient updates). Second, gradient-based methods require the model to be differentiable, which excludes many promising non-differentiable architectures. For example, in attention network design, using FAISS ~\cite{johnson2019faiss} for approximate nearest neighbor hashing to retrieve context from the key-value (KV) cache assumes differentiability, which prevents the use of this approach to implement efficient sparse attention mechanisms.

Igel et al.~\cite{igel2003neuroevolution} and Tim Salimans et al.~\cite{salimans2017evolution} explore using Evolution Strategies (ES) as an alternative to Markov Decision Process (MDP)-based methods for optimizing neural networks in reinforcement learning. In recent years, some methods have utilized EAs to optimize large models' prompts~\cite{sun2022bbtv2,sun2022black,zhao2023genetic,guo2023connecting} and small parameter spaces~\cite{jin2024derivative,carmona2024well,huang2025evolution}, validating the ability of EAs to optimize large model parameters in smaller-scale settings. More recently, Xin Qiu et al.~\cite{qiu2025evolution} adapted the reward computation from GRPO ~\cite{guo2025deepseek} to design the fitness evaluation and population update mechanism in their ES algorithm, successfully extending ES to full-parameter fine-tuning of LLMs in the reinforcement learning stage for the first time. However, these works only apply ES within the reinforcement learning context, and their effectiveness heavily depends on the design of the RL reward function.

We further investigate the capability of Evolution Strategies for direct neural network optimization beyond reinforcement learning. We propose EA4LLM, a method that leverages evolutionary algorithms to optimize LLMs, and demonstrate for the first time that ES can effectively perform full-parameter optimization of LLMs starting from pretraining—validated across model sizes from 0.5B to 32B—as illustrated in \ref{fig:fig1_loss}. In our implementation, we introduce a simple yet effective approach that links the model’s output logits to the fitness computation in ES. This implies that virtually any neural network with a probabilistic training objective $P$ can, in principle, be optimized using ES. Different from prior ES in supervised learning that typically treats mini-batch loss or performance as a black-box objective, we explicitly aggregate token-level log-probabilities (log-softmax) into the ES fitness and carry out full-parameter optimization for LLMs in pretraining; combined with antithetic sampling, rank shaping, grouped/layered controls, and cross-process synchronized subsampled evaluation, this substantially improves the practicality and efficiency of zeroth-order optimization in this setting. Building upon standard ES~\cite{salimans2017evolution}, we conduct extensive exploratory experiments and provide key insights—as well as identified shortcomings—regarding the direct application of ES to neural network optimization.

Our work challenges the assumption that gradient-based methods are the sole viable option for training neural networks. It also holds significant potential to reduce the computational cost of training large language models, thereby enabling research groups with limited computational resources—even those without access to GPUs—to conduct meaningful deep learning research.







\section{Related Work}
\label{sec:background}

Evolutionary Algorithms (EAs) form a vast and diverse family of optimization techniques inspired by the process of natural selection~\cite{back1996evolutionary,zhou2011multiobjective,zhang2008rm}. Among EAs, Evolution Strategies (ES)~\cite{64e965483fda6d7f0630ed47}, which we utilize in our work, constitute merely a small fraction of this extensive domain. ES methods focus on optimizing parameters through random perturbations, crossover, mutation, and other mechanisms to sample a set of solutions. These solutions (individuals) are then evaluated based on their fitness, and the population is updated accordingly. This process repeats until a termination condition is met, such as reaching a maximum number of iterations. Since ES does not require gradient information, they are particularly suitable for non-differentiable problems or scenarios where gradient computation is impractical~\cite{hansen2003reducing}.

Despite the significant potential of using EAs for neural network optimization, research in this area has been relatively limited compared to the widespread adoption of gradient-based methods. Early efforts by Igel et al.~\cite{igel2003neuroevolution} and Tim Salimans et al.~\cite{salimans2017evolution} demonstrated that ES could serve as an effective alternative to traditional reinforcement learning approaches, paving the way for subsequent studies~\cite{lorenc2025utilizingnoveltybasedevolutionstrategies,schulman2017proximalpolicyoptimizationalgorithms}. In recent years, some methods have utilized EAs to optimize large models' prompts~\cite{sun2022bbtv2,sun2022black,zhao2023genetic,guo2023connecting} and small parameter spaces~\cite{jin2024derivative,carmona2024well,huang2025evolution}, validating the ability of EAs to optimize large model parameters in smaller-scale settings. More recently, Xin Qiu et al.~\cite{qiu2025evolution} extended the application of ES to fine-tuning large language models (LLMs) in reinforcement learning, highlighting the algorithm's capability to handle complex, high-dimensional optimization landscapes.

However, these pioneering works primarily focused on specific contexts such as reinforcement learning, leaving ample room for exploring the broader applicability of ES in direct neural network optimization. Our work builds upon this foundation, aiming to unlock the full potential of evolutionary strategies for training neural networks beyond the confines of reinforcement learning. By doing so, we not only challenge the prevailing paradigm of gradient-based optimization but also open new avenues for efficient, resource-conscious deep learning research.

\section{Method}
\label{sec:approach}

\subsection{Linking Logits to ES Fitness}
\label{subsec:logit_fitness}
We optimize the model by connecting its output logits to an Evolution Strategies (ES) fitness function that measures next-token predictive quality on a text corpus.

Let a tokenized sequence be $x = (x_1, x_2, \dots, x_T)$ with vocabulary $\mathcal{V}$. Given parameters $\theta$, the model produces logits $z_t \in \mathbb{R}^{|\mathcal{V}|}$ at position $t$. We convert logits to probabilities via softmax:
\begin{equation}
    p_\theta(x_{t+1} \mid x_{\le t}) 
    = \mathrm{softmax}(z_t)[x_{t+1}] 
    = \frac{\exp\big(z_{t, x_{t+1}}\big)}{\sum_{v \in \mathcal{V}} \exp\big(z_{t, v}\big)}.
\end{equation}

To handle padding and truncation, we introduce a binary mask $m_t \in \{0,1\}$ that excludes pad positions. The per-sample reward (fitness contribution) is the mean log-probability of the true next token:
\begin{equation}
    R(x; \theta) = \frac{1}{\sum_{t=1}^{T-1} m_t} \sum_{t=1}^{T-1} m_t \, \log p_\theta(x_{t+1} \mid x_{\le t}).
\end{equation}

Given a dataset $\mathcal{D}$, we define the overall fitness as the average across samples:
\begin{equation}
    F(\theta) = \frac{1}{|\mathcal{D}|} \sum_{x \in \mathcal{D}} R(x; \theta).
\end{equation}

In ES, a perturbation $\epsilon$ is sampled and applied to parameters as $\theta' = \theta + \sigma \, \epsilon$, yielding a perturbed fitness
\begin{equation}
    r(\epsilon) \triangleq F(\theta + \sigma \, \epsilon) 
    = \frac{1}{|\mathcal{D}|} \sum_{x \in \mathcal{D}} R\big(x; \theta + \sigma \, \epsilon\big),
\end{equation}
which directly ties the model's logits---through $p_\theta$---to the ES objective. In our implementation, we evaluate $R(x;\theta)$ via log-softmax of shifted logits and label-gathering, and average rewards across the dataset, consistent with next-token causal language modeling.

\subsection{Evolution Strategies and Our Extensions}
\label{subsec:es_operator}
\paragraph{Baseline ES Operator.} We adopt a standard ES estimator~\cite{salimans2017evolution}. Let $\epsilon_j \sim \mathcal{N}(0, I)$ for $j = 1, \dots, N$ denote i.i.d. Gaussian noise samples. We evaluate perturbed fitness values
\begin{equation}
    r_j = F\big(\theta + \sigma \, \epsilon_j\big).
\end{equation}
We build normalized weights $w_j$ from $r_j$ to form a gradient-like update. We support either z-score normalization
\begin{equation}
\begin{aligned}
    w_j^{\text{z}} &= \frac{r_j - \mu_r}{\sigma_r + 10^{-8}}, \\
    \mu_r &= \frac{1}{N} \sum_{k=1}^N r_k, \\
    \sigma_r^2 &= \frac{1}{N} \sum_{k=1}^N (r_k - \mu_r)^2
\end{aligned}
\end{equation}
or rank-based shaping
\begin{equation}
    w_j^{\text{rank}} = 2 \cdot \frac{\mathrm{rank}(r_j)}{N-1} - 1 \in [-1, 1],
\end{equation}
where $\mathrm{rank}(\cdot)$ assigns ranks $0,\dots,N-1$ after sorting.

The ES update then aggregates noise weighted by $w_j$:
\begin{equation}
    \hat{g} = \frac{1}{N} \sum_{j=1}^N w_j \, \epsilon_j.
\end{equation}
We apply either the standard scaling or the NES-style $1/\sigma$ scaling:
\begin{equation}
    \theta \leftarrow \theta + \alpha \, \hat{g} \quad \text{(standard)},
    \qquad
    \theta \leftarrow \theta + \frac{\alpha}{\sigma} \, \hat{g} \quad \text{(NES with $1/\sigma$)}.
\end{equation}


\paragraph{Subsampled Evaluation.} To reduce per-iteration computation, we evaluate fitness on a random subset of the dataset instead of all samples. At iteration $t$, let $\mathcal{D}_t \subset \mathcal{D}$ be a uniformly sampled subset with size $M$ (without replacement). We define the mini-batch fitness
\begin{equation}
    F_M(\theta) = \frac{1}{M} \sum_{x \in \mathcal{D}_t} R(x; \theta), \qquad M \ll |\mathcal{D}|.
\end{equation}
Perturbed evaluations use the same subset:
\begin{equation}
    r_j^{(t)} = F_M\big(\theta + \sigma \, \epsilon_j\big), \qquad j = 1, \dots, N.
\end{equation}
This yields a stochastic ES estimator that is unbiased under uniform sampling, i.e., $\mathbb{E}_{\mathcal{D}_t}[F_M(\theta)] = F(\theta)$, while substantially reducing wall-clock time and memory. In our implementation, the subset indices are broadcast across processes so all workers evaluate the same $\mathcal{D}_t$, and antithetic sampling plus rank shaping further mitigates the variance introduced by sub-sampling.

\begin{algorithm}[tb]
   \caption{EA4LLM: Zeroth-Order Pretraining via ES}
   \label{alg:basic_es}
\begin{algorithmic}
   \STATE {\bfseries Input:} Pretrained LLM parameters $\theta_0$; token-level fitness $R(x;\theta)$ via log-softmax; iterations $T$; population size $N$; noise scale $\sigma$; learning rate $\alpha$; subset size $M$; normalization (z-score or rank shaping).
   \FOR{$t = 1$ {\bfseries to} $T$}
      \STATE Sample a uniform subset $\mathcal{D}_t \subset \mathcal{D}$ with $|\mathcal{D}_t|=M$ and broadcast indices across processes.
      \STATE Draw $\{\epsilon_j\}_{j=1}^N$ with $\epsilon_j \sim \mathcal{N}(0,I)$.
      \FOR{$j = 1$ {\bfseries to} $N$}
         \STATE Form temporary parameters: $\theta_{\text{tmp}} = \theta + \sigma\,\epsilon_j$.
         \STATE Evaluate reward on the subset: $r_j = \frac{1}{M} \sum_{x \in \mathcal{D}_t} R\big(x; \theta_{\text{tmp}}\big)$.
      \ENDFOR
      \STATE Compute weights $w_j$ from $\{r_j\}$ by z-score normalization or rank shaping.
      \STATE Aggregate noise: $\hat{g} = \frac{1}{N} \sum_{j=1}^N w_j\, \epsilon_j$.
      \STATE Update parameters: $\theta \leftarrow \theta + \alpha\, \hat{g}$.
   \ENDFOR
   \STATE {\bfseries Return:} $\theta_T$.
\end{algorithmic}
\end{algorithm}

The pseudocode of our final algorithm is shown in Algorithm~\ref{alg:basic_es}.

\section{Experiment}
\label{sec:experiment}

\subsection{Experimental Setups}

\paragraph{Global Hyperparameters and Setup.} Unless otherwise stated, we adopt the following configuration drawn from our experiment runner and training script: 
\begin{itemize}
    \item \textbf{Population}: $P=30$; \textbf{noise scale}: $\sigma=10^{-3}$; \textbf{step size}: $\alpha=5\times10^{-4}$.
    \item \textbf{Iterations}: $1000$ with $5000$ evaluation samples per iteration; seeds are synchronized across processes.
    \item \textbf{Micro-batch}: $16$; \textbf{max sequence length}: $256$;
    \item \textbf{Models}: Qwen3 configurations spanning 0.5B--32B (see Appendix~\ref{sec:appendix_model_config}).
    \item \textbf{Data sampling}: up to $10{,}000$ lines from a novel corpus.
    \item \textbf{Baseline toggles}: antithetic off, $1/\sigma$ scaling off, rank shaping off, grouped multipliers off; scheduler decays disabled ($\gamma_\alpha=\gamma_\sigma=1.0$, $S_\alpha=S_\sigma=0$).
\end{itemize}

\subsection{Experimental Results}

\paragraph{Cross-Scale Training Validation (0.5B--32B).} We pretrain Qwen3-style models from 0.5B to 32B with full-parameter updates and consistently observe stable loss reduction; the trend matches the loss curve shown earlier in Figure~\ref{fig:fig1_loss}, indicating ES effectively optimizes across model scales.

\begin{figure}[t]
  \centering
  \includegraphics[width=\linewidth]{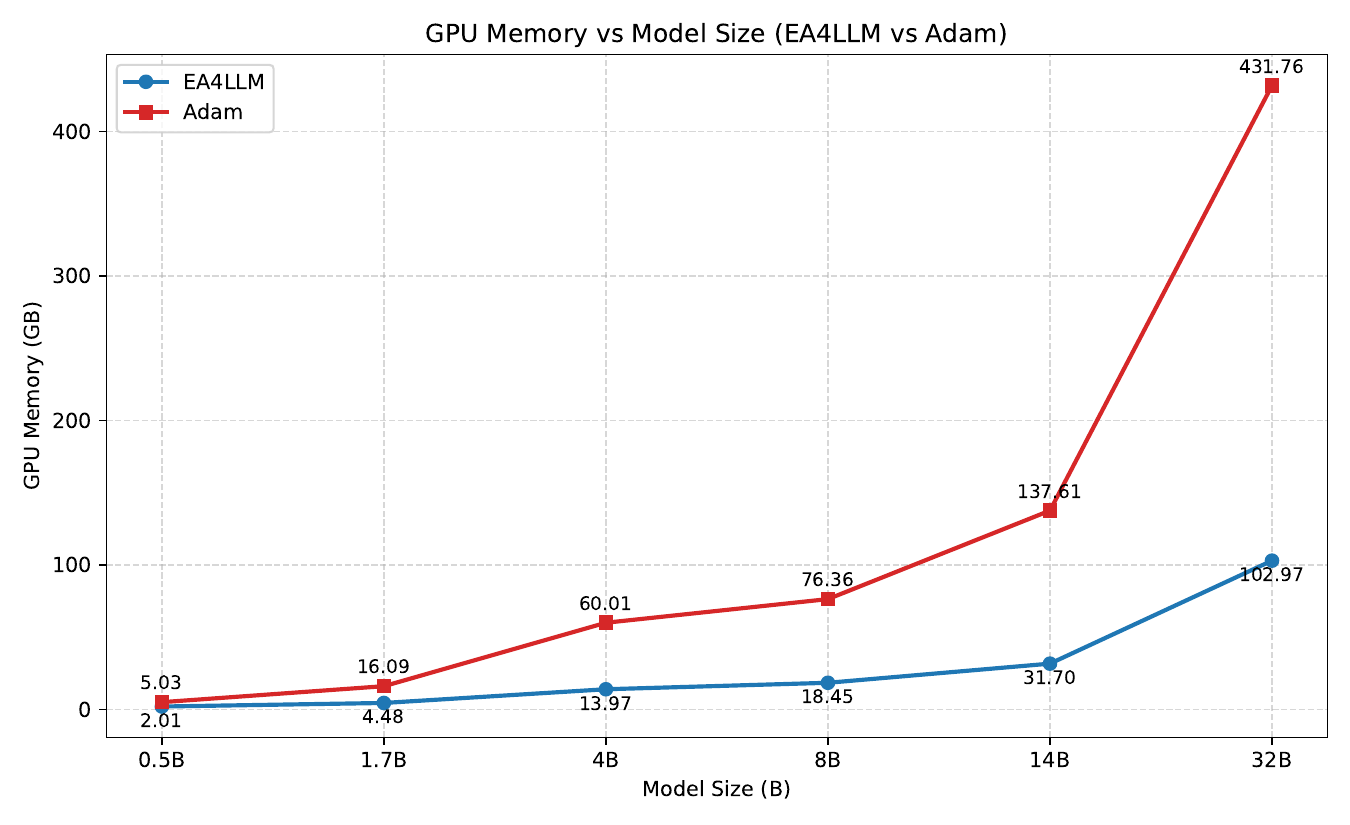}
  \caption{Minimal GPU memory usage across model sizes (0.5B--32B), comparing ES with gradient-based optimization. ES consistently uses less memory, with a larger margin as the model size increases.}
  \label{fig:memory_exp2_plot}
\end{figure}

\paragraph{Memory Footprint Comparison: Evolutionary vs Gradient Optimization.} We compare the minimal GPU memory usage when training 0.5B--32B models using ES versus gradient-based pipelines. By avoiding backpropagation and the storage of gradients/optimizer states, ES exhibits lower memory usage across all scales; the relative advantage increases with model size. Moreover, as model parameters increase, the memory advantage becomes more pronounced; for models of 4B and above, ES-based pretraining reduces GPU memory usage by more than $4\times$ compared to gradient-based methods. See Figure~\ref{fig:memory_exp2_plot}.

\begin{figure}[t]
  \centering
  \includegraphics[width=\linewidth]{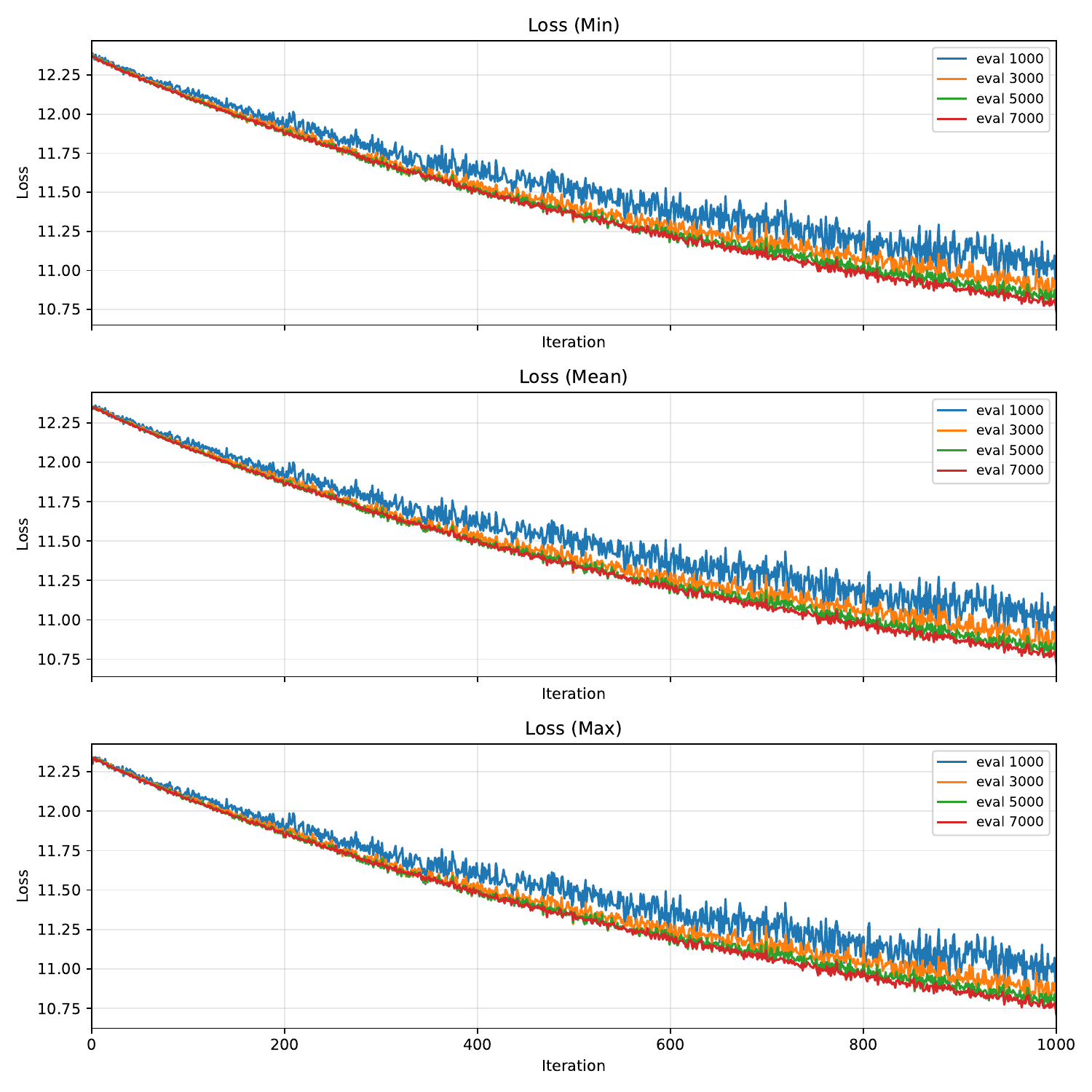}
  \caption{Different evaluation budgets (1.7B). Larger budgets yield more accurate optimization signals and stronger loss reduction.}
  \label{fig:ablation_eval_samples}
\end{figure}

\begin{figure}[t]
  \centering
  \includegraphics[width=\linewidth]{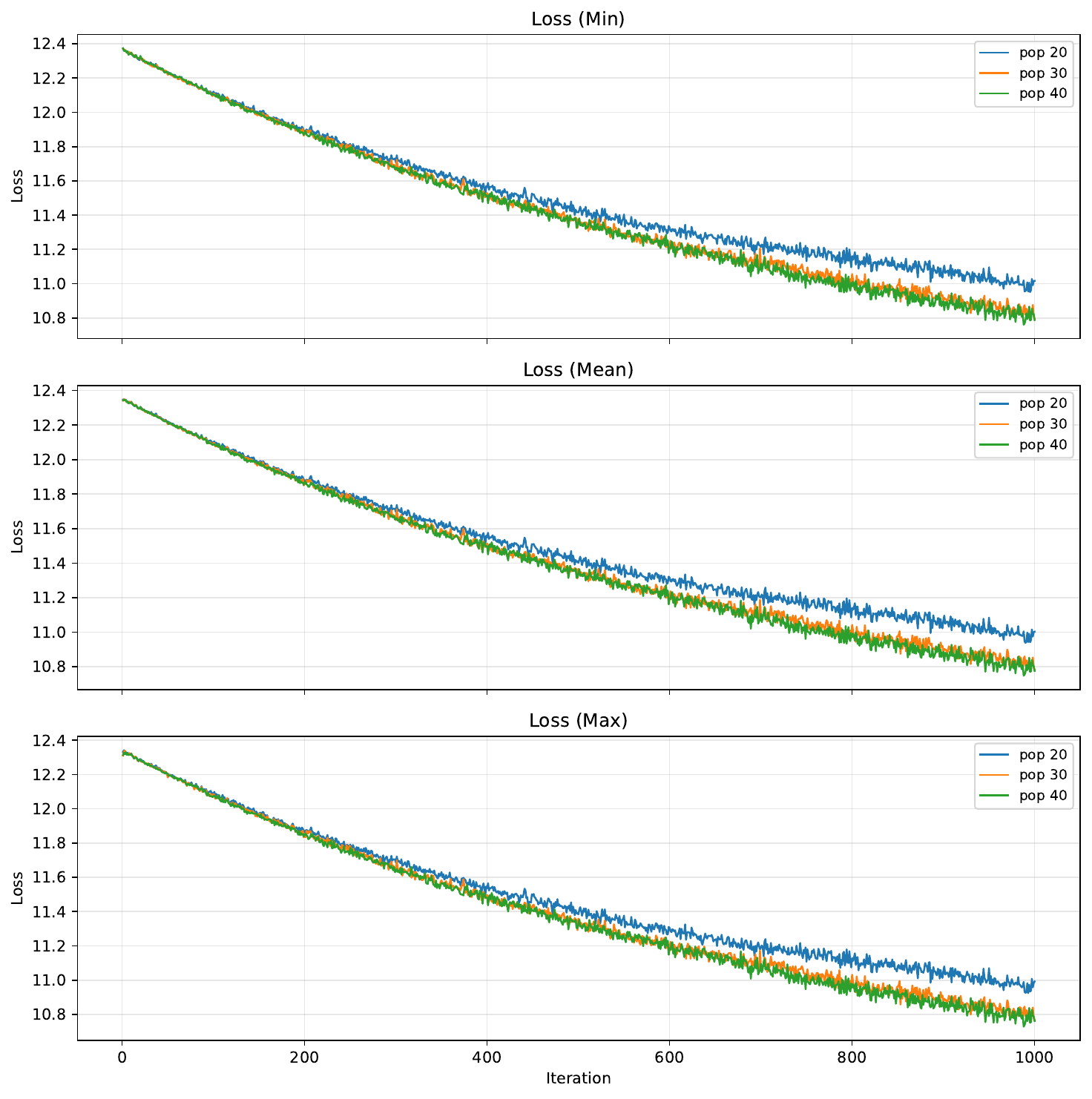}
  \caption{Different population sizes (eval=5000). Larger populations yield more accurate optimization signals and stronger loss reduction.}
  \label{fig:ablation_population}
\end{figure}

\paragraph{Ablations: Evaluation Budget and Population Size.} On the 1.7B model, we study the effect of evaluation samples per iteration (1000/3000/5000/7000) and population size (20/30/40). Results match expectations: larger evaluation budgets and larger populations provide more accurate optimization signals, leading to more pronounced loss reduction. Considering time and performance constraints, in other experiments we primarily adopt evaluation budget 5000 and population size 30. See Figure~\ref{fig:ablation_eval_samples} and Figure~\ref{fig:ablation_population}.

\section{Conclusion}

\paragraph{The Significance of Our Work}
Our approach provides a practical, architecture-agnostic zeroth-order pretraining path for large language models by tying token-level log-softmax to an ES fitness (Section~\ref{sec:approach}). We employ subsampled evaluation to reduce per-iteration cost while maintaining an unbiased estimator under uniform sampling, and aggregate perturbations via weights from either z-score or rank shaping. Experimentally (Section~\ref{sec:experiment}), ES delivers stable loss reduction from 0.5B to 32B with full-parameter updates, and achieves substantial memory advantages over gradient-based training: for models of 4B and above, ES-based pretraining reduces GPU memory usage by more than $4\times$. Ablations on a 1.7B model confirm expected trends: larger evaluation budgets and populations provide more accurate optimization signals and stronger loss reduction; in practice, we primarily adopt an evaluation budget of 5000 and a population of 30.

\paragraph{Limitations and Future Work}
While promising, zeroth-order ES pretraining has limitations. Variance can be higher than first-order gradients and performance is sensitive to population size $N$, noise scale $\sigma$, learning rate $\alpha$, and normalization choice; subsampling, though unbiased, introduces additional stochasticity. Sample efficiency may trail gradient-based methods because each iteration requires multiple forward evaluations. Scaling to extremely large models depends on efficient parameter perturbation and IO. In our experiments, certain variance-reduction and control mechanisms (e.g., antithetic sampling, NES-style $1/\sigma$ scaling, grouped multipliers) were disabled; exploring these and hybrid pipelines (alternating ES and gradient steps) is a natural direction. Future work includes adaptive schedules for $\alpha$ and $\sigma$, improved variance reduction, and task-aware fitness shaping beyond next-token log-likelihood to better align with downstream objectives.

\nocite{langley00}

\bibliography{main}
\bibliographystyle{icml2025}

\newpage
\appendix
\onecolumn
\section*{Appendix}

\appendix

\section{Comparative Analysis: Differences and Significance}
\label{sec:appendix_comparison}

\subsection{Differences from ES in Supervised Learning}
Salimans et al. (2017) studied, on small supervised datasets (e.g., MNIST), the correlation between ES gradient estimates \(g_t^{ES}\) obtained via random perturbations and the true gradients \(g_t^{True}\) from backpropagation, using the mini-batch performance/loss as a black-box objective \(f(\theta+\sigma\epsilon)\). While that objective is computed from logits, it treats the loss as an opaque function and does not offer a full-parameter ES optimization of large-scale LLMs for pretraining.

Our approach explicitly reformulates the maximum-likelihood language modeling objective as ES fitness. Let the model outputs (logits) be \(z_t\). We define
\begin{equation}
    R(x;\theta) = \frac{1}{T}\sum_{t=1}^T m_t \cdot \log \, \mathrm{softmax}\big(z_t(\theta)\big)_{y_t}, \quad F(\theta) = \frac{1}{|\mathcal{D}|}\sum_{x\in\mathcal{D}} R(x;\theta),
\end{equation}
where \(m_t\) can be used for masking or weighting, and \(F(\theta)\) aligns with the standard log-likelihood objective in language modeling. We use a simple population ES with i.i.d. Gaussian perturbations \(\epsilon_j\sim\mathcal{N}(0,I)\); for each iteration we evaluate \(F(\theta + \sigma\epsilon_j)\), apply either z-score normalization or rank-based shaping to the rewards, estimate the ES direction, and update parameters with the core ES step \(\theta \leftarrow \theta + \alpha \, \hat g\). This design removes antithetic sampling, grouped/layered operators, and step-size multipliers, matching the actual algorithm we employ for large-scale pretraining while remaining effective empirically.

\subsection{Differences from ES in the RL Stage}
RL-stage ES and GRPO-style methods focus on task-specific rewards for fine-tuning and depend on reward design and sampling environments. In contrast, we apply ES \emph{from pretraining} to full-parameter optimization with fitness defined by the aggregation of token-level log-probabilities consistent with maximum likelihood, independent of external environments or rewards. This directly targets the pretraining objective and avoids reliance on RL-specific reward shaping.

\subsection{Differences from MeZO and Other Zeroth-Order Optimizers}
MeZO is a memory-efficient zeroth-order optimizer that typically uses two function evaluations \(f(\theta\pm\delta u)\) to approximate directional derivatives and then updates parameters in a first-order style, with the objective treated as a black-box mini-batch loss. In contrast, our method:
\begin{itemize}
    \item uses \emph{population-based} ES with i.i.d. Gaussian perturbations, without antithetic pairs or grouped/layered operators;
    \item integrates the \emph{structured token-level log-softmax} objective into ES fitness, rather than treating the loss as opaque;
    \item stabilizes updates via \emph{z-score} or \emph{rank-based} reward normalization to reduce scale sensitivity and outlier effects;
    \item demonstrates \emph{cross-scale effectiveness} (0.5B--32B) and a \emph{memory footprint reduction} of over 4\(\times\) for models \(\geq\)4B during pretraining; in other experiments we primarily adopt evaluation budget 5000 and population size 30.
\end{itemize}
Therefore, our method directly optimizes the structured probabilistic objective of language modeling with a simple ES update and practical normalization strategies, making zeroth-order pretraining viable at LLM scale.


\subsection{Model Configuration Details}
\label{sec:appendix_model_config}
We instantiate a series of Qwen3-style causal language models using the Transformers implementation (\texttt{Qwen3Config}) spanning 0.5B--32B parameters in our experiments. The key configuration settings for each size are summarized below.

\begin{table}[t]
\centering
\small
\setlength{\tabcolsep}{4pt}
\begin{tabular}{lrrrrrrrrrr}
\hline
Size & Layers & Hidden & Heads & KV Heads & Head Dim & Intermediate & Max Pos & Tie Emb & Use Cache & Vocab \\
\hline
0.5B & 24 & 1024 & 8 & 4 & 128 & 4096 & 32768 & true & false & 151936 \\
1.7B & 28 & 2048 & 16 & 8 & 128 & 6144 & 32768 & true & true & 151936 \\
4B & 36 & 4096 & 32 & 8 & 128 & 9728 & 32768 & true & true & 151936 \\
8B & 36 & 4096 & 32 & 8 & 128 & 12288 & 32768 & false & true & 151936 \\
14B & 40 & 5120 & 40 & 8 & 128 & 17408 & 32768 & false & true & 151936 \\
32B & 64 & 8192 & 64 & 8 & 128 & 25600 & 40960 & false & true & 151936 \\
\hline
\end{tabular}
\caption{Qwen3-style model configurations used for ES pretraining experiments across 0.5B--32B.}
\label{tab:model_configs}
\end{table}


\end{document}